\documentclass[journal]{IEEEtran}
\usepackage{cite}
\usepackage{amsmath,amssymb,amsfonts}
\usepackage{algorithmic}
\usepackage{graphicx}
\usepackage{hyperref}
\usepackage{textcomp}
\usepackage{multirow}
\usepackage{adjustbox}
\begin{document}
\title{SSL-CPCD: Self-supervised learning with composite pretext-class discrimination for improved generalisability in endoscopic image analysis}
\author{Ziang Xu, Jens Rittscher, and
       Sharib Ali
\thanks{Z. Xu and J. Rittscher are with the Institute of Biomedical Engineering, Department of Engineering Science, University of Oxford, OX3 7DQ, Oxford, United Kingdom (emails: \{ziang.xu, jens.rittscher\} @eng.ox.ac.uk)}
\thanks{S. Ali is with School of Computing, Faculty of Engineering and Physical Sciences, University of Leeds, LS2 9JT, Leeds, United Kingdom (corresponding email: s.s.ali@leeds.ac.uk)}
}

\maketitle
\begin{abstract}
Data-driven methods have shown tremendous progress in medical image analysis. In this context, deep learning-based supervised methods are widely popular. However, they require a large amount of training data and face issues in generalisability to unseen datasets that hinder clinical translation. Endoscopic imaging data incorporates large inter- and intra-patient variability that makes these models more challenging to learn representative features for downstream tasks. Thus, despite the publicly available datasets and datasets that can be generated within hospitals, most supervised models still underperform. While self-supervised learning has addressed this problem to some extent in natural scene data, there is a considerable performance gap in the medical image domain. In this paper, we propose to explore patch-level instance-group discrimination and penalisation of inter-class variation using additive angular margin within the cosine similarity metrics. Our novel approach enables models to learn to cluster similar representative patches, thereby improving their ability to provide better separation between different classes. Our results demonstrate significant improvement on all metrics over the state-of-the-art (SOTA) methods on the test set from the same and diverse datasets. We evaluated our approach for classification, detection, and segmentation. SSL-CPCD achieves 79.77\% on Top 1 accuracy for ulcerative colitis classification, 88.62\% on mAP for polyp detection, and 82.32\% on dice similarity coefficient for segmentation tasks are nearly over 4\%, 2\%, and 3\%, respectively, compared to the baseline architectures. We also demonstrate that our method generalises better than all SOTA methods to unseen datasets, reporting nearly 7\% improvement in our generalisability assessment.
\end{abstract}


\section{Introduction}
\label{sec:introduction}
\IEEEPARstart{I}{mage} classification, detection, and segmentation tasks have been extensively studied by the biomedical image analysis community~\cite{LITJENS201760}. 
Recent advances in data-driven approaches are mostly based on convolutional neural networks (CNNs) and have gained interest due to their ability to surpass traditional machine learning approaches. CNNs have been widely used for multiple tasks and different imaging modalities, including computed tomography (CT)~\cite{kaul2019focusnet}, X-ray~\cite{Azizi2021BigSM}, magnetic resonance imaging (MRI)~\cite{chen2019self} and endoscopy~\cite{AliNPJ_review22}. 

Supervised learning-based approaches in machine learning (ML) are data-voracious and not generalisable on out-of-distribution datasets. Obtaining labelled data is a significant hurdle for medical image analysis as it requires clinical expertise. Additionally, it accounts for the risk of human bias proportional to the sample size~\cite{lux2013annotation}. Data curation challenges are thus harder to tackle, leading only to sub-optimal results in supervised learning frameworks~\cite{zhou2021preservational}. Several studies have also found that most supervised methods lead to a huge performance drop when applied to different centre datasets~\cite {zhou2021preservational}. Changes in patient population, the appearance of lesions, imaging modalities used, and differences in hardware all affect data variability, pose a bottleneck during training, and adversely affect model performance. We ask if we can leverage already available high-quality public datasets with and without labels to fine-tune these models on a new, small, and out-of-distribution dataset without compromising algorithmic performance but instead boosting them. 

Self-supervised learning (SSL) learns more semantically meaningful features by training a ML model using unlabelled data first which is then fine-tuned on a smaller training sample with the available labelled samples for each specific downstream task, thus eliminating the requirement of a large amount of labelled data during training improving generalisation capability for the next downstream task and expansion to other out-of-distribution datasets~\cite{huang2021towards}. In the medical imaging field, SSL has been used extensively for different tasks, including disease classification~\cite{Azizi2021BigSM,zhuang2019self}, lesion region detection~\cite{chen2019self,nguyen2020self} and segmentation~\cite{zeng2019sese,ciga2022self}.

\begin{figure}[t!]
    \centering
    \includegraphics[width=0.5\textwidth]{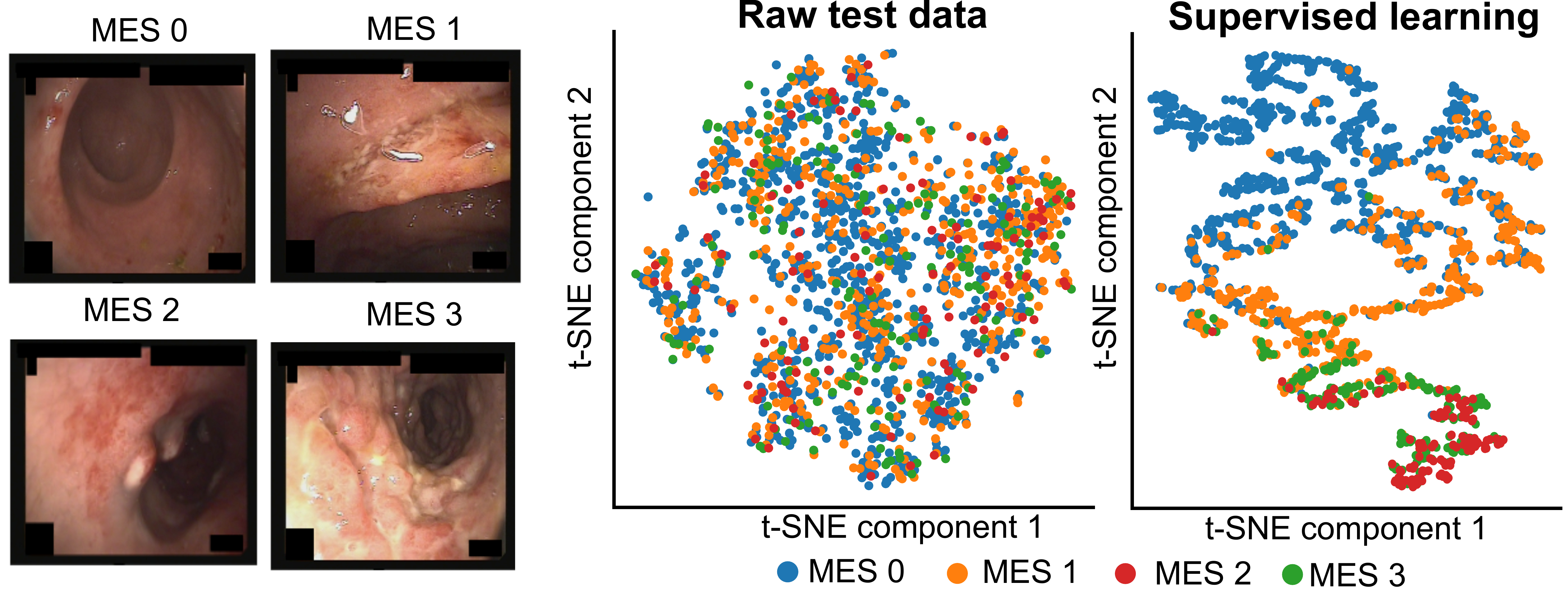}
    \caption{Endoscopic image analysis for ulcerative colitis scoring. (on left) Representative images for Mayo endoscopic scoring (MES) from 0 to 3, and (on right) t-SNE plots for all test samples before learning and after supervised learning using ResNet50. \label{fig:motivation}}
    \end{figure}

Endoscopy remains the clinical standard for diagnosing and surveying disease in hollow organs. In contrast to data obtained from other imaging modalities, the analysis of endoscopy video is extremely challenging~\cite{AliNPJ_review22} due to various factors such as internal organ deformation, light interaction with tissue at different depths, imaging artefacts such as bubbles, fluid and other floating objects, and a considerable operator dependency. Subtle and fine-grained changes often indicate the onset of disease. The robust computer-aided techniques of such changes poses a significant challenge. In this work, we focus on two different lesions found in the colon and rectum, and we aim to devise a robust SSL-based approach to build automated techniques with CNN-based networks. To this end, we propose to develop SSL approaches for ulcerative colitis (UC), a chronic intestinal inflammatory disease, and polyps that precursor lesions for colorectal cancer.
Gastroenterologists use the Mayo Endoscopic Score (MES, see Fig.~\ref{fig:motivation} (on the left)), a widely accepted predictive indicator for malignant transformation in UC, as a classification task based on visual appearances. Similarly for polypectomy (aka removal of polyps) is also based on visual cues. Automated classification, detection and segmentation methods can therefore help reduce missed operator variability in these procedures. 

Supervised learning methods struggle to learn a feature representation that discriminates between the different categories even if trained on large, labelled datasets. The data presented in Fig.~\ref{fig:motivation} illustrates this problem in the context of ulcerative colitis scoring. After supervised learning, we can still observe significant confusion between the different classes. 
In this work, we propose a novel self-supervised learning strategy for endoscopic image analysis, referred to as ``SSL-CPCD''. Our approach is based on novel ideas on combining loss functions both at the single instance-level and group-level instance using image frames and patch-level representations. The proposed losses are used in a pretext-invariant representation learning context where patch-level and image-level representations are learnt at single and clustered group-level instances, amplifying the power of learning discriminative features and are framed as a Noise Contrastive Estimation (NCE) function. Jointly, we refer to this loss as a composite pretext-class discrimination loss (CPCD). Our novel approach improves learning on the pretext task by using both image-level and patch-level discrimination. To this extent, we also use memory banks to store positive and negative samples with moving weights that help to learn features that are semantically meaningful for downstream tasks. 
In addition, we use penalisation of inter-class variation between positive and negative samples using additive angular margin in our instance-level contrastive loss. 
By transforming images into jigsaw puzzles and computing contrastive losses between different feature embeddings, we learn a representation capable of differentiating between the subtle characteristics of the different classes.

In addition, we also explore the introduction of an attention mechanism in our network for further improvement. 
Key contributions of our presented work can be summarised below:
\begin{itemize}
    \item Novel SSL-CPCD method can learn semantically meaningful features from a large amount of unlabeled data, leading to improved performance on subsequent tasks, including classification, detection, and segmentation of two different lesion types.
    \item Single and group-level instances are used to minimise noise contrastive estimation loss to increase the inter-class separation and minimise the intra-class distance. 
    \item Inclusion of an additive angular margin within the cosine similarity metric in the contrastive loss to further penalise decision boundary with respect to the negative samples further increases inter-class separation.
    \item Evaluation of our method on four different datasets including Kvasir-SEG~\cite{jha2020kvasir}, CVC-ClinicDB~\cite{bernal2015wm}, LIMUC~\cite{polat2022labeled}, and our in-house dataset. 
    \item We show that our SSL-CPCD-based method outperforms several SOTA SSL strategies by a large margin.
\end{itemize}

\section{Related work}
\subsection{Deep learning in gastrointestinal endoscopy}
\subsubsection{Classification task}
Ulcerative colitis (UC) scoring in clinic is based on Mayo Endoscopic Scoring (MES). CNN-based methods were used in efforts for automating these scoing. For example, Stidham et al.~\cite{stidham2019performance} used an Inception V3 model to train and evaluate MES scores in still endoscopic frames where they used 16k UC images and obtained an accuracy of 67.6\%, 64.3\% and 67.9\% for the three MES classes.
Recently, Mokter et al.~\cite{mokter2020classification} proposed a method to classify UC severity in colonoscopy videos by detecting vascular (vein) patterns using three CNN networks and using a training dataset comprising over 67k frames. Similarly, Ozawa et al.~\cite{ozawa2019novel} used a CNN for binary classification only to elevate the problem of poor accuracies across classes and used still frames comprising 26k training images, which first between normal (comprising of MES 0 and MES 1) while next class as combined moderate (MES 2) and severe (MES 3). Gutierrez et al.~\cite{becker2021training} also used the CNN model to predict only a binary version of the MES scoring.
\subsubsection{Detection task}
Polyp detection task has been more widely researched compared to UC classification. Lee et al.~\cite{lee2020real} used YOLOv2 and validated the algorithm on public datasets and colonoscopy videos demonstrating real-time capability as one of the milestone. Zhang et al.~\cite{zhang2019real} proposed Single Shot MultiBox Detector (SSD) for gastric polyps. They linked the feature maps from the lower layers, and the feature maps deconvolved from the upper layers and improved the mean precision (mAP) from 88.5\% to 90.4\%. Qadir et al.~\cite{qadir2019polyp}, and Shin et al.~\cite{shin2018automatic} used Mask R-CNN and Faster RCNN with different backbones to detect polyps, respectively. Although high precision is obtained but limited in processing speed.
\subsubsection{Segmentation task}
Polyp segmentation task is the most widely researched topic in the endoscopic image analysis.
%
Zhou et al.~\cite{zhou2019unet++} proposed a technique called U-Net++ based on U-Net, which fully utilises multi-scale features to obtain superior results. Fan et al.~\cite{fan2020pranet} proposed a parallel inverse attention based network (PraNet). PraNet employs a partial decoder to aggregate features in high-level layers, and mine boundary cues using an inverse attention module. A Shallow Attention Network (SANet) was proposed by~\cite{wei2021shallow}. SANet used color swap operation to decouple image content and color, and force the model to pay more attention to the shape and structure of the object.
Recently, Srivastava et al.~\cite{srivastava2021msrf} proposed a Multi-Scale Residual Fusion Network (MSRF-Net). MSRF-Net can exchange multi-scale features of different receptive fields using dual-scale dense fusion (DSDF) blocks. 
%
\subsection{Attention mechanism}
Attention can make the model more focused, extract the most relevant features, and ignore irrelevant information. It also overcomes the size limitation of the receptive field and can focus on the contribution of global features to the current region~\cite{jaderberg2015spatial}. Attention-based models have achieved state-of-the-art performance in medical images such as skin cancer, endoscopy, CT, and X-ray (Sinha and Dolz~\cite{sinha2020multi}, Zhao et al.~\cite{zhao2021adasan}, Kaul et al.~\cite{kaul2019focusnet}, Gu et al.~\cite{gu2020net}). Zhao et al.~\cite{zhao2021adasan} proposed an adaptive cosine similarity network with a self-attention module to automatically classify gastrointestinal endoscope images. The self-attention block replaces the conv+BN/Relu operation in traditional CNN and uses cosine-based self-adapting loss function to adjust the scale parameters automatically achieving 95.7\% on average accuracy in the wireless capsule endoscopy dataset. 
%
%
\begin{figure*}[h!t!]
    \centering
    \includegraphics[width=0.92\textwidth]{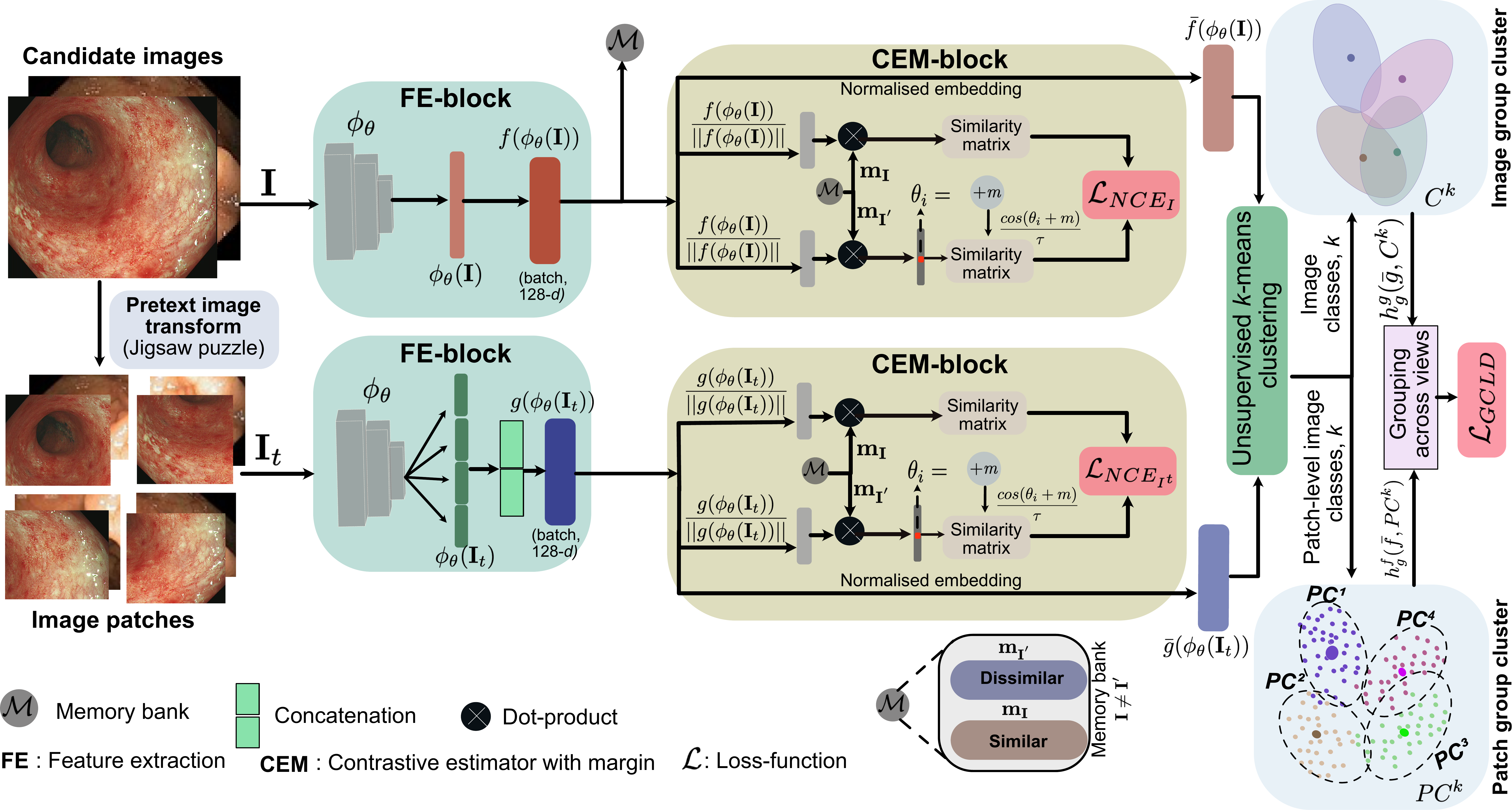}
    \caption{\textbf{Block diagram of our proposed self-supervised learning framework with composite pretext-class discrimination losses (SSL-CPCD).} ResNet50 encoder network is fine-tuned with original images with transformations and jigsaw puzzle patches in a self-supervised setting to enable semantically meaningful representation learning for improved generalisability and accuracy in downstream tasks. Contrastive estimator with margin (CEM-block) is separately computed for image-level and path-level instances. Further, a group-wise contrastive loss is computed by comparing the centroids at patch-level (PC$^k$) group and at image-level (C$^k$). Memory bank $\mathcal{M}$ is used for storing all representations.}
    \label{fig:my_label}
\end{figure*}
\subsection{Self-supervised learning}
Self-supervised learning (SSL) uses pretext tasks to mine self-supervised information from large-scale unsupervised data, thereby learning valuable image representations for downstream tasks. By doing so, the limitation of network performance on predefined annotations is greatly reduced. In SSL, the pretext task typically applies a transformation to the input image and predicts the properties of the transformation from the transformed image. Chen {\it et al.}~\cite{chen2020simple} proposed the SimCLR model, which performs data enhancement on the input image to simulate the input from different perspectives of the image. A contrastive loss is then used to maximise the similarity of the same object under different data augmentations and minimise the similarity between similar objects. Later, the MoCo model proposed by He et al.~\cite{he2020momentum} also used contrastive loss to compare the similarity between a query and the keys of a queue to learn feature representation. The authors used a dynamic memory, rather than a static memory bank, to store feature vectors used in training.
In contrast to these methods that encourage the construction of covariant image representations to the transformations, pretext-invariant representation learning (PIRL)~\cite{misra2020self} pushes the representations to be invariant under image transformations. PIRL computes high similarity to the image representations similar to the representation of the transformed versions and low resemblance to representations for the different images. The notion of a Jigsaw puzzle~\cite{noroozi2016unsupervised} was used as a pretext task for PIRL representation learning. 

In recent years, self-supervised learning has also been applied in the field of medical image analysis but not much on the endoscopic image analysis. Azizi et al.~\cite{Azizi2021BigSM} used multi-instance contrastive learning based on self-supervision on medical images, followed by a fully supervised fine-tuning method for the final classification of available task-specific losses. They improved top-1 accuracy by 6.7\% and 1.1\% on dermatology and chest X-ray classification, respectively. Zeng et al.~\cite{zeng2019sese} proposed SeSe-Net for medical image segmentation. SeSe-Net is divided into two neural networks, "worker" and "supervisor". In the first stage, the standard data set is learned and segmented, and a training set is generated, and then the supervisor further supervises the learning process in the second stage so that the worker further improves the performance on the non-labelled dataset. Chen et al.~\cite{chen2019self} proposed a novel self-supervised learning strategy based on context restoration to change the spatial information of an image by selecting and exchanging two patches in the same image to learn enough pronounced semantic representations. It was validated on 2D fetal ultrasound images, abdominal computed tomography images, and brain magnetic resonance images. Recently, Ciga et al.~\cite{ciga2022self} used a residual network pre-trained with self-supervised learning to learn generalisable features and then used the pre-trained network in downstream tasks to perform multiple tasks on multiple multi-organ digital histopathology datasets. 
\section{Methodology}
We propose a novel self-supervised approach that exploits the invariant representation learning beneficial for downstream tasks by using both image-level instance-group discrimination and patch-level instance-group discrimination losses. To this extent, we propose two novel approaches - firstly, exploiting positive and negative samples for noise contrastive loss estimation ($\mathcal{L}_\text{NCE}$). Unlike classically used $\mathcal{L}_\text{NCE}$~\cite{gutmann2010noise}, we integrate an added angular margin between the negative embeddings and learned normalised weights. This enables to dissociation of different samples further. Secondly, we apply group-wise cross-view associations between patches and images. Our novel group-wise loss enables us to learn fine-grained features at both patch-level $\mathcal{L}_\text{PC}^k$ and image level $\mathcal{L}_\text{C}^k$  that can enhance more local representations. For grouping of the embeddings, here we utilise a $k$-means clustering technique with class numbers similar to downstream tasks to provide representative clusters. Our approach uses memory banks to store all representations useful for various loss function estimations. Below we have described each element of our approach presented in the block diagram in Fig.~\ref{fig:my_label}.
\subsection{Feature extraction (FE) block}{\label{sec:FE}}
Let the endoscopy dataset $D$, which consist $N$ represents image samples denoted as $\mathcal{D}=\{\textbf{I}_1, \textbf{I}_2, ..., \textbf{I}_N\}$. We use a transformation $\mathcal{T}$ to create and reshuffle $m$ number of image patches for each image in each dataset($\mathcal{D}$, $\mathcal{P}=\{\textbf{I}_{1t}^{1}, ...,\textbf{I}_{1t}^{m}, ..., \textbf{I}_{Nt}^{1}, ..., \textbf{I}_{Nt}^{m}\}$ with $\mathcal{T}\in t$). We train a convolutional neural network (\textit{in our case}, ResNet50) with free parameters $\theta$ that embody representation $\phi_\theta(\textbf{I})$ for a given sample $\textbf{I}$ and $\phi_\theta(\textbf{I}_t)$ for patches $\mathcal{P}$.

\noindent
\textbf{Image-level embedding}: Candidate images are fed in batches which are transformed using simple geometric (horizontal and vertical flips) and photometric (colour jitter with 0.4 for hue, saturation, contrast and brightness) transformations and fed into an encoder giving a feature representation $\phi_\theta(\textbf{I})$. We then apply a projection head $f(.)$ to re-scale the representations to a $128$-dimensional feature vector.

\noindent
\textbf{Patch-level embedding}: For image patches, representations of each patch constituting the image $\textbf{I}$ are concatenated to form $\phi_\theta(\textbf{I}_t)$. A projection head $g(.)$ is applied to re-scale the representations to a $128$-dimensional feature vector. However, in this case, we perform random cropping and shuffle of the cropped areas into patch size of $64\times 64$ along with the colour transforms used for the original images. 

\noindent
\textbf{Memory banks}: The memory bank $\mathcal{M}$ stores all the feature representations of the dataset $\mathcal{D}$ at the image level computed from the original images $\textbf{I}$. These embedding weights are moving average of feature representation $f(\phi_\theta(\textbf{I}))$ represented as $\textbf{m}_\textbf{I}$ with assigned indexes that helps to build negative samples $\textbf{m}_{\textbf{I}^{'}}$ for each image during contrastive loss estimation. $\mathcal{M}$ is updated at every epoch with the step-size of 0.5 $\times$ initial weight and normalised to between 0 and 1 similar to~\cite{misra2020self}.

\subsection{Contrastive loss estimation with margin  (CEM) block}{\label{sec:CEM}}
Noise contrastive estimator (NCE)~\cite{pmlr-v9-gutmann10a,misra2020self} is used to measure the similarity scores $s$. In our noise contrastive estimator, a positive sample pair (\textbf{I}, $\textbf{I}_t$) has $n^{-}$ corresponding ``negative samples'', \textit{i.e.,} representations of each sample other than $\textbf{I}$ ($\textbf{I}^{'}$). Moving average representations for the positive and negative samples ($\textbf{m}_\textbf{I}$, $\textbf{m}_{\textbf{I}^{'}}$) are used from the memory bank to perform a dot-product between the normalised target feature embedding $t^{+}$ and the normalised positive sample representations $\textbf{m}_\textbf{I}$ (\textit{i.e., a cosine similarity}). However, unlike~\cite{pmlr-v9-gutmann10a,misra2020self}, we propose to add an angular margin to increase the separation between the target embedding $t^{+}$ and the ``negative samples'' $\textbf{m}_{\textbf{I}^{{'}}}$ in our contrastive loss estimation with margin block (CEM-block). We do this by first computing angular separation between the positive target embedding and negative embedding ($\psi$) $\psi = \arccos<t^{+}, \textbf{m}_{\textbf{I}^{{'}}}>$ and then add an angular margin $m$ to the computed angle, i.e. $\psi_{new} = \psi + m$. Finally, $cosine$ of the $\psi_{new}$ gives our new similarity between the positive and negative samples. The same CE block is applied for both image-level and patch-level NCE loss computations. The NCE models the probability of the binary event given a data distribution $\mathcal{D}$ and temperature parameter $\tau$:
\begin{align}{\label{eq:nce11}}
   \! h(\phi_\theta(\textbf{I}),\phi_\theta(\textbf{I}_t))\!\!  =  \!\! \frac{\exp\frac{{<\phi_\theta(\textbf{I}), \phi_\theta(\textbf{I}_t)>}}{\tau}}{\exp\frac{{<\phi_\theta(\textbf{I}), \phi_\theta(\textbf{I}_t)>}}{\tau} + \!\! \sum_{\textbf{I}^{'}\in n^{-}}\!\! \exp\frac{<\phi_\theta(\textbf{I}^{'}), \phi_\theta(\textbf{I}_t)>}{\tau}}
\end{align}
Adding the angular additive margin to further penalise dissimilarity between target embedding and negative samples, the above equation can be rewritten as:
\begin{align}{\label{eq:nce12}}
    h^{'}(\phi_\theta(\textbf{I}),\phi_\theta(\textbf{I}_t)) =  \frac{\exp\frac{{<\phi_\theta(\textbf{I}), \phi_\theta(\textbf{I}_t)>}}{\tau}}{\exp\frac{{<\phi_\theta(\textbf{I}), \phi_\theta(\textbf{I}_t)>}}{\tau} + \frac{\text{cos}(\phi_{new})}{\tau}},\\
   \text{with}~ \phi_{new} = \arccos(\exp(<\phi_\theta(\textbf{I}^{'}), \phi_\theta(\textbf{I}_t)>)+m) 
\end{align}
The total NCE loss entails to minimise the joint-loss function both at the image-level and patch-level configurations:
\begin{align}{\label{eq:NCE_total}}
    \mathcal{L}_{NCE}^{total}(\textbf{I},\textbf{I}_t) = \lambda \mathcal{L}_{{NCE}_{I}}(\textbf{m}_{\textbf{I}}, f(\phi_\theta(\textbf{I}))) + \\ \nonumber
    (1-\lambda) \mathcal{L}_{{NCE}_{{I}^t}}(\textbf{m}_{\textbf{I}}, g(\phi_\theta(\textbf{I}^t)))
\end{align}
with each loss component can be established as a joint-probability distribution between the target embedding, positive and negative embeddings given as:
\begin{align}{\label{eq:loss_each}}
\mathcal{L}_{{NCE}_{I}}(\textbf{m}_{\textbf{I}}, f(\phi_\theta(\textbf{I}))) = -\log [h(f(\phi_\theta(\textbf{I})), \textbf{m}_{\textbf{I}})]    \nonumber  \\
   -\sum_{\textbf{I}'\in \mathcal{D}_n} \log [1- h( \textbf{m}_{\textbf{I}^{'}}, f(\phi_\theta(\textbf{I})))], \text{and} \\
   \mathcal{L}_{{NCE}_{I^t}}(\textbf{m}_{\textbf{I}}, g(\phi_\theta(\textbf{I}^t))) = -\log [h(g(\phi_\theta(\textbf{I}^t)), \textbf{m}_{\textbf{I}})]    \nonumber  \\
   -\sum_{\textbf{I}'\in \mathcal{D}_n} \log [1- h( \textbf{m}_{\textbf{I}^{'}}, g(\phi_\theta(\textbf{I}^t)))].
\end{align}
The configured joint-loss $\mathcal{L}_{NCE}^{total}(\textbf{I},\textbf{I}_t)$ enables to learn representations of image \textbf{I} closer to its transformed counterpart $\textbf{I}_t$ and also to the memory representation $\textbf{m}_{\textbf{I}}$ that will damp the parameter updates in the weights $\phi_{\theta}$. It also further penalises the representations from other set of images $\textbf{I}^{'}$.
\subsection{$k$-means feature grouping}
One important limitation of single instance discrimination as done in NCE loss is that they focus on within-instance similarity by data augmentation assuming a single distinctive instance, but in downstream tasks, these can appear as various similar observations of the same instance. Thus, a grouping strategy can help mitigate such limitations, as presented in this Section.
\textbf{Normalised projection head:} We utilise the linear projection heads to normalise the feature embedding with $l_2$-norm that enables to reduce variance from data augmentation and maps the features onto a unit hypersphere, $\bar{f}(\phi_\theta(\textbf{I})) = \frac{f(\phi_\theta(\textbf{I}))}{||{f}(\phi_\theta(\textbf{I}))||}$, and $\bar{g}(\phi_\theta(\textbf{I}_t)) = \frac{g(\phi_\theta(\textbf{I}_t))}{||{g}(\phi_\theta(\textbf{I}_t))||}$.\\

\textbf{Feature grouping:} To overcome the limitation of the single instance approach, we have used grouping instances based on the local clusters within a batch of samples. We create $k$ clusters where $k$ is the number of classes (say $n$) in the downstream tasks and use this to define clusters at image and patch levels. Using spherical $k$-means clustering, we group the unit-length feature vectors. We compute the cluster centroids for each image embedding $C^k$ and patch embedding $PC^k$ in batch input with $k=\{1, ..., n\}$, where $n$ is the number of cluster classes depending on the downstream task. We assign each instance in the image and patches to each of their corresponding nearest centroids, say $C(i) = j$, meaning instance $i$ is assigned to centroid $j$ and so on.
\subsection{Cross-level discrimination at image and patch-levels}{\label{sec:CLD}}
\textbf{Cross-level grouping:} Clusters could be noisy, so we applied a cross-view local group for each instance by an element-wise multiplication of the feature embedding at image-level $\bar{f}(\phi_\theta(\textbf{I}))$ with the cluster centroid of image patches $PC^k{_i}$, and at patch-level $\bar{g}(\phi_\theta(\textbf{I}))$ with the cluster centroid $C^k_{i}$ of the images in the batch where $i$ is the feature embedding assigned to the cluster. 

\textbf{Cross-level contrastive loss:} The noise contrastive estimation (NCE) loss across the views can be defined using a very similar expression as in Eq.~\eqref{eq:nce11}. However, here, we will use the group cluster embeddings and centroids, and we want to assume that the group in the patch-level cluster is identical to the group in image level for that specific class. Thus, our cross-level contrastive loss at group-level can be defined as:
\begin{align}{\label{eq:nce13}}
    \!\!h_g^f(\bar{f}_i, PC_i)\!\!= \!\! -\log \frac{\exp\frac{{<\bar{f}_i, PC_i>}}{\tau}}{\exp\frac{<\bar{f}_i, PC_i>}{\tau} + \!\! \sum_{j\neq i}{\exp\frac{<\bar{f}_i, PC_j>}{\tau}}}
\end{align}
Similarly, cross-level grouping of the patch-level representations can be written as:
\begin{align}{\label{eq:nce14}}
    h_g^g(\bar{g}_i, C_i)\!\!= \!-\log \frac{\exp\frac{{<\bar{g}_i, C_i>}}{\tau}}{\exp\frac{<\bar{g}_i, C_i>}{\tau} + \!\! \sum_{j\neq i}{\exp\frac{<\bar{g}_i, C_j>}{\tau}}}
\end{align}
The final group-wise cross-level discrimination loss $\mathcal{L}_{GCLD}$ incorporating both the image-level and the patch-level representations combined with equal weights ($\lambda=0.5$) can be written as:
\begin{align}{\label{eq:GCLD_loss}}
   \!\!\mathcal{L}_{GCLD}\!\!  =\!\! \sum_{k=1}^{k=4}\sum_{i=1}^{N} \{\lambda h_g^f(\bar{f}_i, PC_i^k) + (1-\lambda)  h_g^g(\bar{g}_i, C_i^k)\}
\end{align}
\subsection{Proposed CPCD loss}
Our final novel loss function that defines single instance-level, group-level, and cross-level representations as a joint loss optimisation problem is referred to as \textit{composite pretext-class discrimination loss} (CPCD). In this work, the contrastive noise loss referring to single instance-level representations and the group-wise cross-level discrimination loss are equally weighted (say, $\lambda^{'} = 0.5$. Thus, the final CPCD loss $\mathcal{L}_{CPCD}$ is given as:
\begin{align}
   \!\! \mathcal{L}_{CPCD} = \sum_{k=1}^{k=4} \lambda^{'}\underbrace{ \sum_{i=1}^{N} \{\lambda h_g^f(\bar{f}_i, PC_i^k) + (1-\lambda)  h_g^g(\bar{g}_i, C_i^k)\}}_{\mathcal{L}_{GCLD}} +  \nonumber \\ \underbrace{(1-\lambda^{'}) \sum_{i=1}^{N}
   \{ - \lambda \log(\textbf{m}_{\textbf{I}}, f(\phi_\theta(\textbf{I})))
    -\log (1-\lambda)  (\textbf{m}_{\textbf{I}}, g(\phi_\theta(\textbf{I}^t)))\}}_{\mathcal{L}_{NCE}}
\end{align}
\section{Experiments}
\subsection{Dataset and setup}
\begin{table} [t!]
 \caption{{Colonoscopic datasets used in our experiments}}
    \label{table:datasettable}
   \footnotesize
    \centering
    \begin{tabular}{p{2.5cm}|p{0.8cm}|p{1.2cm}|p{0.5cm}|p{0.5cm}|p{0.5cm}}
        \hline
        \textbf{Dataset} & \textbf{Images} & \textbf{Input size} & \textbf{Train} & \textbf{Valid} & \textbf{Test}\\ 
        \hline
        \multicolumn{5}{l}{\textbf{Ulcerative colitis classification}}\\
        \hline
        LIMUC~\cite{polat2022labeled} & 11276 & $224\times 224$ & 8631 &959 & 1686\\ \hline
        In-house & 251 & $224 \times 224$ &0 &0 &251 \\  \hline
        \textbf{Polyp segmentation}\\
        \hline
        Kvasir-SEG~\cite{jha2020kvasir}  & 1000 & Variable &800 &100 &100\\  \hline
        SUN(non-polyp)~\cite{misawa2021development}  & 1000 & Variable &1000 &0 &0\\  \hline
        CVC-ClinicDB~\cite{bernal2015wm} & 612 & $384\times 288$ &0  & 0& 612 \\  \hline
        \textbf{Polyp detection}\\
        \hline
        Kvasir-SEG~\cite{jha2020kvasir}  & 1000 & Variable &800 &100 &100\\  \hline
        SUN~\cite{misawa2021development}  & 1000 & Variable &1000 &0 &0\\  \hline
\end{tabular}
\end{table}	
\begin{table}[!t]
\centering
\footnotesize
\caption{$k$-fold cross-validation on baseline ResNet50}
\begin{tabular}{@{}l|l|l|l|l|l|l@{}}
\hline
\textit{k} & \textbf{Top 1} & \textbf{Top 2} & \textbf{F1-score} & \textbf{Specificity} & \textbf{Recall} & \textbf{QWK}\\ 
\hline
\hline
3  & 0.7361	&0.8991	&0.6542 & 0.8519	&0.6732	&0.8099 \\ \hline
5  & \textbf{0.7539}	&\textbf{0.9277}	& \textbf{0.6702}	& \textbf{0.8670}	&0.6906	&\textbf{0.8251}\\ \hline
10 & 0.7532	&0.9198	&0.6689	&0.8649	&\textbf{0.6932}	&0.8239\\ \hline
\end{tabular}
\label{tab:result3}
\end{table}
\begin{table*}[!ht]
\centering
\footnotesize
\caption{Quantitative comparison for UC classification task on LIMUC dataset}
\begin{tabular}{@{}l|l|l|l|l|l|l|l@{}}
\hline
\textbf{Method} & \textbf{Backbone} & \textbf{Top 1} & \textbf{Top 2} & \textbf{F1} & \textbf{Spec.} & \textbf{Recall} & \textbf{QWK}\\ 
\hline
\hline
Baseline & R50  & 0.7539 & 0.9277 & 0.6702 & 0.8670 & 0.6906 & 0.8251 \\ \hline
Baseline & R50-Att.   & 0.7562 & 0.9591 & 0.6742 & 0.8715	& 0.7035 & 0.8278\\ \hline
SimCLR~\cite{chen2020simple} & R50 & 0.7355 & 0.9387 & 0.6631 & 0.8510 & 0.6752 & 0.8083 \\ \hline
SimCLR~\cite{chen2020simple} & R50-Att.   & 0.7384 & 0.9219 & 0.6649 & 0.8431 & 0.6942 & 0.8102 \\ \hline
SimCLR+DCL~\cite{chuang2020debiased} & R50 & 0.7555 & 0.9450 & 0.6729 & 0.8635 & 0.6897 & 0.8269 \\ \hline
SimCLR+DCL~\cite{chuang2020debiased} & R50-Att.   & 0.7568 & 0.9367 & 0.6755 & 0.8669 & 0.6952 & 0.8287 \\ \hline
MoCoV2+CLD~\cite{he2020momentum} & R50 & 0.7574 & 0.9493 & 0.6788 & 0.8721 & 0.6959 & 0.8309 \\ \hline
MoCoV2+CLD~\cite{he2020momentum} & R50-Att.   & 0.7598 & 0.9536 & 0.6812 & 0.8709 & 0.7047 & 0.8333 \\ \hline
PIRL~\cite{misra2020self} & R50 & 0.7651 & 0.9637 & 0.6859 & 0.8874 & 0.7098 & 0.8376 \\ \hline
PIRL~\cite{misra2020self} & R50-Att.   & 0.7740 & 0.9610 & 0.6918 & 0.8893 & 0.7133 & 0.8460  \\ \hline
SSL-CPCD (ours) & R50 & 0.7912 & 0.9633 & 0.7209 &	\textbf{0.9043} & 0.7198 & 0.8693 \\ \hline
SSL-CPCD (ours) & R50-Att.   & \textbf{0.7977} & \textbf{0.9750}	& \textbf{0.7279} & 0.9008 & \textbf{0.7259} & \textbf{0.8746} \\ \hline
\end{tabular}
\label{tab:result1}
\end{table*}
\subsubsection{Dataset}
We have explored various colonoscopic imaging datasets that are available publicly and in-house for three different downstream tasks. For the classification task, LIMUC~\cite{polat2022labeled} and one in-house dataset (collected under universal patient consenting at the Translational Gastroenterology Unit, John Radcliffe Hospital, Oxford) are applied. Kvasir-SEG~\cite{jha2020kvasir} and CVC-ClinicDB~\cite{bernal2015wm} are used for the segmentation task. Similarly, Kvasir-SEG~\cite{jha2020kvasir} for experiments on polyp detection as a downstream task. In the pretext task for the detection and segmentation tasks, we have used polyp samples from Kvasir-SEG~\cite{jha2020kvasir} and non-polyp samples from the SUN~\cite{misawa2021development} dataset for training our SSL model. The details about the datasets and the number of training, validation, and testing samples used are presented in Table~\ref{table:datasettable}. 
\subsubsection{Evaluation metrics}
We have used standard top-$k$ accuracy (percentage of samples predicted correctly, top1 and top2 are used), F1-score ($=\frac{2tp}{2tp+fp+fn}$, tp: true positive, fp: false positive), specificity ($=\frac{tp}{tp+fn}$), sensitivity or recall ($=\frac{tn}{tn+fp}$), and Quadratic Weighted Kappa (QWK) for our classification task. For the detection task, standard computer vision metrics, including mean average precision (mAP at an IoU interval [0.25:0.05:0.75]) and AP small, medium and large, were used for our experiments. Dice similarity coefficient (DSC), which is also known as F1-score, and type-II error referred to as F2-score, recall and positive predictive values (PPV, $=\frac{tp}{tp+fp}$) have been used for evaluating our segmentation task.
\subsubsection{Implementation details} 
The proposed method is implemented using PyTorch~\cite{paszke2019pytorch}. All experiments were conducted on an NVIDIA Quadro RTX 6000 graphics card. For pretext tasks in self-supervised learning, we have used the batch size of 32 and trained our model for 2000 epochs in all experiments or until convergence with stopping criteria. The SGD optimiser with a learning rate of $1e^{-3}$ was used for training. All input images were resized to $224\times224$ pixels.

For the downstream classification task, we have fine-tuned the model with a learning rate of 1$e^{-4}$, the SGD optimiser with a batch size of 32, and the learning rate decay of 0.9 times per 20 epochs. In Table~\ref{tab:result3}, we tested the effect of k-fold cross-validation on model training for different k-value settings. The experimental results show that the model achieves the highest Top-1 accuracy and QWK when k = 5. Therefore, we use 5-fold cross-validation in our experiments. For the detection task, we have used the Adam optimiser with a learning rate of 1$e^{-5}$ and a batch size of 32 to finetune for 400 epochs. For the segmentation task, 300 epochs with a batch size of 16 and an SGD optimiser with a learning rate of 1$e^{-3}$ were used to finetune the model. All experiments used 80\% of the dataset for training, 10\% for validation, and the remaining held-out 10\% for testing. We additionally have used out-of-centre unseen centre datasets for generalisability study.
\textbf{Hyperparameters:} 
For group-wise cross-level discrimination loss ($\mathcal{L}_{GCLD}$ in Eq.~(\ref{eq:GCLD_loss})), we set $k=4$ for a number of clusters in classification pretext task, $k=2$ in detection and segmentation pretext task, $s=6$ for the re-scaling and $m=0.5$ for an angular margin. 
Memory bank proposed in~\cite{misra2020self} has been used with the same hyperparameters. For Eq.~(\ref{eq:NCE_total}) we use $\lambda = 0.5$ and use $\tau = 0.4$ for computing the function $h(.,.)$ in Eq.~(\ref{eq:nce11}, \ref{eq:nce12}, \ref{eq:nce13}, and \ref{eq:nce14}). We used an updated weight of 0.5 for the memory bank exponential moving average representations. These values are justified in our ablation study provided in Section~\ref{sec:ablation}. 
\begin{table*}[!htb]
\centering
\footnotesize
\caption{Quantitative comparison for polyp detection task using Kvasir-SEG dataset}
\begin{tabular}{@{}l|l|l|l|l|l|l|l|l@{}}
\hline
\textbf{Method} & \textbf{Backbone} & \textbf{mAP} & \textbf{AP25} & \textbf{AP50} & \textbf{AP75 } & \textbf{APsmall} & \textbf{APmedium} & \textbf{APlarge}\\ 
\hline
\hline
RetinaNet~\cite{lin2017focal} & R50  &0.8637	&0.9377	&0.8965	&0.6973	&0.4832	&0.7507	&0.8398 \\ \hline
RetinaNet~\cite{lin2017focal} & R50-Att.  &0.8729 &0.9436	 &0.9097	&0.7045 &0.4871	&0.7603 &0.8419\\ \hline
SimCLR~\cite{chen2020simple} & R50 & 0.8501	&0.9259	&0.8837	&0.6709	&0.4641	&0.7416	&0.8237\\ \hline
SimCLR~\cite{chen2020simple} & R50-Att.   & 0.8532 &0.9278	&0.8818	&0.6846	&0.4679	&0.7403	&0.8302\\ \hline
SimCLR+DCL~\cite{chuang2020debiased} & R50 & 0.8537	&0.9269	&0.8853	&0.6852	&0.4709	&0.7429	&0.8321\\ \hline
SimCLR+DCL~\cite{chuang2020debiased} & R50-Att.   & 0.8592	&0.929	&0.8893	&0.6887	&0.4739	&0.7467	&0.8403 \\ \hline
MoCoV2+CLD~\cite{he2020momentum} & R50 & 0.8505	&0.9273	&0.9028	&0.6845	&0.4779	&0.7491	&0.8346 \\ \hline
MoCoV2+CLD~\cite{he2020momentum} & R50-Att.  & 0.8633	&0.9412	&0.9044	&0.7019	&0.4859	&0.7563	&0.8402 \\ \hline
PIRL~\cite{misra2020self} & R50 & 0.8612	&0.9403	&0.8931	&0.6929	&0.4839	&0.7487	&0.8317 \\ \hline
PIRL~\cite{misra2020self} & R50-Att.   & 0.8677	&0.9408	&0.8961	&0.702	&0.4863	&0.7589	&0.8408  \\ \hline
SSL-CPCD (ours) & R50 & 0.8709	&0.9421	&0.9192	&0.7107	&0.5033	&0.763	&0.8542 \\ \hline
SSL-CPCD (ours) & R50-Att.   & \textbf{0.8862}	&\textbf{0.9469}	&\textbf{0.9227}	&\textbf{0.7197}	&\textbf{0.5105}	&\textbf{0.7703}	&\textbf{0.8598} \\ \hline
\end{tabular}
\label{tab:result4}
\end{table*}
\subsection{Results}
In this section, we present the comparison of our proposed SSL-CPCD approach with other SOTA SSL methods.
\subsubsection{Comparison for UC classification task}
ResNet50 (R50) and ResNet50 with convolution-block attention module (R50-Att.) are established as the baseline model for supervised learning first, and then the same are used for other SOTA SSL-based method comparisons in Table~\ref{tab:result1} for ulcerative colitis classification task on LIMUC dataset. Baseline networks R50 and R50-Att., respectively, obtained 75.39\% and 75.62\% on top-1 accuracy, and 82.51\% and 82.78\% on QWK. Our proposed SSL-CPCD method yielded the best results with 79.77\%, 72.79\%, 90.08\%, 72.59\% and 87.46\% on top 1 accuracy, F1 score, specificity, recall and QWK, respectively. Compared to the supervised learning-based baseline models (R50), the top 1 accuracy and QWK is improved by 4.38\% and 4.95\%, respectively, using our proposed SSL-CPCD with the same backbones. We also compared our proposed SSL-CPCD approach with other SOTA SSL methods, including popular SimCLR~\cite{chen2020simple}, SimCLR+DCL~\cite{chuang2020debiased}, MoCoV2+CLD~\cite{he2020momentum} and PIRL~\cite{misra2020self} methods. Our proposed network (R50-Att.) outperformed all these methods with at least nearly 2.4\% (PIRL) up to 6\% (SimCLR) on top-1 accuracy. Similar improvements can be observed on other metrics as well. 
\subsubsection{Comparison for polyp detection task}
The Kvasir-SEG polyp dataset was used to evaluate the performances of SSL on detection as the downstream task in endoscopy. The quantitative results from Table~\ref{tab:result4} show that our proposed SSL-CPCD approach outperforms all the other SOTA methods on all metrics. It achieves 2.29\%, 2.7\% and 3.3\% improvement on mAP compared to SSL methods, including MoCoV2+CLD, SimCLR+DCL and SimCLR, respectively. Our method also improves 1.83\% on AP50 and 1.4\% on APmedium (medium polyp sizes) compared to MoCoV2+CLD, respectively. Compared to the widely used supervised technique RetinaNet, our method is better on mean average mAP but significantly improves over AP50, AP75 and size-based metrics.
\begin{table}[!tb]
\centering
\footnotesize
\caption{Quantitative comparison for polyp segmentation task}
\begin{tabular}{@{}l|l|l|l|l|l@{}}
\hline
\textbf{Method} & \textbf{Backbone} & \textbf{DSC} & \textbf{F2} & \textbf{Recall} & \textbf{PPV}\\ 
\hline
\hline
U-Net~\cite{ronneberger2015u} & none  & 0.7933	&0.7671	&0.7945	&0.9131\\ \hline
Res-UNet~\cite{zhang2018road} & R50  & 0.7867	&0.7667	&0.7723	&0.9139 \\ \hline
Res-UNet~\cite{zhang2018road} & R50-Att.  & 0.792	&0.7743	&0.7862	&0.9187\\ \hline
SimCLR~\cite{chen2020simple} & R50 & 0.7892	&0.7621	&0.7639	&0.9146\\ \hline
SimCLR~\cite{chen2020simple} & R50-Att.   & 0.7945	&0.7759	&0.7903	&0.9162\\ \hline
SimCLR+DCL~\cite{chuang2020debiased} & R50 & 0.7879	&0.7609	&0.7653	&0.9169\\ \hline
SimCLR+DCL\cite{chuang2020debiased} & R50-Att.   & 0.7933	&0.7822	&0.7741	&0.9038 \\ \hline
MoCoV2+CLD\cite{he2020momentum} & R50 & 0.7946	&0.779	&0.7846	&0.9173 \\ \hline
MoCoV2+CLD\cite{he2020momentum} &R50-Att.   & 0.8029	&0.7953	&0.7998	&0.9201 \\ \hline
PIRL~\cite{misra2020self} & R50 & 0.7906	&0.7842	&0.7946	&0.9135 \\ \hline
PIRL~\cite{misra2020self} & R50-Att.  & 0.7969	&0.7893	&0.8056	&0.9177  \\ \hline
SSL-CPCD (ours) & R50 & 0.8173 & 0.8032 & 0.8104 & 0.9217 \\ \hline
SSL-CPCD (ours) & R50-Att.   & \textbf{0.8232}	& \textbf{0.8081}	&\textbf{0.8234}	&\textbf{0.9259} \\ \hline
\end{tabular}
\label{tab:result5}
\end{table}
%
%
\subsubsection{Comparison for polyp segmentation task}
The Kvasir-SEG dataset was also used to assess the performance of SSL-based approaches in our experiment for segmentation as a downstream task in endoscopy. Table~\ref{tab:result5} compares the result of the proposed SSL-CPCD with other SOTA SSL  approaches and baseline supervised model.  
While proposed SSL-CPCD provided an improvement of 3.12\% and 3.72\% on DSC and Recall, respectively, for the baseline ResNetUNet in a supervised setting, our approach also showed improvements of 2.03\%, 1.28\%, 2.36\% and 0.58\% over MoCoV2 + CLD in DSC, F2-score, recall and PPV, respectively. Higher recall while keeping the precision (PPV) high (over 90\%) indicates that our method is more medically relevant.
\subsection{Generalisation}
To ensure the generalisation of the proposed approach, we trained our model and other methods on one dataset and then tested them on an unseen dataset from different institutions.  
\subsubsection{Generalisibility study for UC classification}
We used the UC classification model trained on the LIMUC dataset collected at Marmara University School of Medicine. We tested this model on our in-house dataset (collected at the John Radcliffe Hospital, Oxford). Table~\ref{tab:result2} assess the generalisability of our SSL-CPCD model and other SOTA approaches on UC classification task. Our proposed SSL-CPCD obtained an acceptable Top 1 accuracy of 67.33\%, F1-score of 64.69\%, specificity of 86.77\%, recall of 64.03\% and QWK of 78.87\%. While outperforming all SOTA approaches, compared with MoCoV2+CLD, our method achieves an improvement of 5.98\% on top 1 accuracy and nearly 9\% in QWK. Table~\ref{tab:result2} shows that our SSL-CPCD outperforms other SOTA methods in various evaluation metrics.
\begin{table}[!tb]
\centering
\footnotesize
\caption{Generalisation study for the UC classification task} 
\begin{tabular}{@{}l|l|l|l|l|l@{}}
\hline
 \textbf{Method}  &  \textbf{Backbone} & \textbf{Top 1} & \textbf{Spec.} & \textbf{Recall} & \textbf{QWK}\\ 
\hline
\hline
 Baseline  & R50  & 0.5856  & 0.7239 & 0.5569 & 0.5379 \\ \hline
 Baseline & R50-Att.   & 0.6055  & 0.7539 & 0.5739 & 0.6572\\ \hline
SimCLR\cite{chen2020simple} & R50 & 0.5737  & 0.7020 & 0.5256 & 0.5611\\ \hline
SimCLR\cite{chen2020simple} & R50-Att.   & 0.5777  & 0.7139 & 0.5420 & 0.6018\\ \hline
SimCLR+DCL\cite{chuang2020debiased} & R50 & 0.5976   & 0.7297 & 0.5622 & 0.6345\\ \hline
SimCLR+DCL\cite{chuang2020debiased} & R50-Att.   & 0.6016   & 0.7458 & 0.5758 & 0.6542 \\ \hline
MoCoV2+CLD\cite{he2020momentum} & R50 & 0.6175  & 0.7716 & 0.5878 & 0.6939 \\ \hline
MoCoV2+CLD\cite{he2020momentum} & R50-Att.   & 0.6135  & 0.7823 & 0.5737 & 0.6902 \\ \hline
PIRL\cite{misra2020self} & R50 & 0.6255  & 0.8213 & 0.6097 & 0.7312 \\ \hline
PIRL\cite{misra2020self}& R50-Att.   & 0.6335  & 0.8397 & 0.6139 & 0.7469  \\ \hline
SSL-CPCD(ours) & R50 & 0.6534  &	0.8501	& 0.6249 & 0.7835 \\ \hline
SSL-CPCD(ours) & R50-Att.   & \textbf{0.6733} & \textbf{0.8677} &	\textbf{0.6403} & \textbf{0.7887} \\ \hline
\end{tabular}
\label{tab:result2}
\end{table}
\subsubsection{Generalisability study for polyp segmentation}
All models for both baseline and SOTA approaches were first trained on the Kvasir-SEG dataset and then tested on the CVC-ClinicDB dataset, for which the results are presented in Table~\ref{tab:result6}. Our proposed SSL-CPCD drastically surpassed baseline supervised approaches (over 10\% on DSC for U-Net and over 7\% on DSC with the same backbone on ResUNet). In addition, our method obtained an improvement of 4.73\% and 6.8\%, respectively, over in MoCoV2+CLD and SimCLR on DSC. Similarly, over 5\% improvement on PIRL is evident in both backbone settings (R50 and R50-Att.). 
%
%
\subsection{Ablation studies}
We have conducted an extensive ablation study of our approach. First, we ablated the impact of multiple loss functions, including NCE, GCLD, and the added angular margin $m$. Then, we conducted an ablation study experiment to further evaluate the performance of our proposed approach under different parameter settings.
\begin{table}[!t]
\centering
\footnotesize
\caption{Generalisation study for segmentation task}
\begin{tabular}{@{}l|l|l|l|l|l@{}}
\hline
\textbf{Method} & \textbf{Backbone} & \textbf{DSC} & \textbf{F2-score} & \textbf{Recall} & \textbf{PPV}\\ 
\hline
\hline
U-Net~\cite{ronneberger2015u} & none  & 0.5826	&0.6029	&0.5942	&0.7633\\ \hline
Res-UNet~\cite{zhang2018road} & R50  & 0.6092	&0.6379	&0.6265	&0.8218 \\ \hline
Res-UNet~\cite{zhang2018road} & R50-Att.   & 0.6027	&0.6499	&0.6372	&0.8065\\ \hline
SimCLR~\cite{chen2020simple} & R50 & 0.5942	&0.6498	&0.6334	&0.8312\\ \hline
SimCLR~\cite{chen2020simple} & R50-Att.   & 0.6113	&0.6556	&0.6673	&0.8329\\ \hline
SimCLR+DCL\cite{chuang2020debiased} & R50 & 0.6039	&0.6501	&0.6586	&0.8293\\ \hline
SimCLR+DCL\cite{chuang2020debiased} & R50-Att.   & 0.6092	&0.6679	&0.6598	&0.8301 \\ \hline
MOCOv2+CLD\cite{he2020momentum} & R50 & 0.6268	&0.6498	&0.6509	&0.8277 \\ \hline
MOCOv2+CLD\cite{he2020momentum} & R50-Att.   & 0.632	&0.6691	&0.6675	&0.8362 \\ \hline
PIRL\cite{misra2020self} & R50 & 0.6196	&0.6742	&0.6703	&0.8378 \\ \hline
PIRL~\cite{misra2020self} & R50-Att.   & 0.6277	&0.6801	&0.6770	&0.8396  \\ \hline
SSL-CPCD (ours)& R50 & 0.6705	&0.6903	&0.6812	&0.8379 \\ \hline
SSL-CPCD (ours)& R50-Att.   & \textbf{0.6793}	&\textbf{0.6978}	&\textbf{0.6897}	&\textbf{0.8488} \\ \hline
%
\end{tabular}
\label{tab:result6}
\end{table}
%
%
%
\begin{table}[!b]
    \centering
    \caption{Ablation study on different loss functions}
    \begin{adjustbox}{width=\columnwidth,center}
    \begin{tabular}{l|c|c|c|c|c|c|c}
    \hline
         \multicolumn{2}{l|}{\textbf{Loss}}& \multicolumn{2}{c|}{\textbf{Class. task}} & \multicolumn{2}{c|}{\textbf{Det. task}}  & \multicolumn{2}{c}{\textbf{Seg. task}}  \\ \cline{3-8}
          \multicolumn{2}{l|}{\textbf{function}} & Top 1 & F1 & AP50 & mAP & DSC & PPV \\ \hline
        \multicolumn{2}{l|}{NCE} & 0.7651 & 0.6859 & 0.8931 & 0.8612 & 0.7906 & 0.9135 \\ \hline
        \multicolumn{2}{l|}{NCE+GCLD} & 0.7858 & 0.7193 & 0.9136 & 0.8683 & 0.8135 & 0.9207 \\ \hline
        \multicolumn{2}{l|}{NCE+GCLD+$m$}& \textbf{0.7912} & \textbf{0.7209} & \textbf{0.9192} & \textbf{0.8709} & \textbf{0.8173} & \textbf{0.9218} \\ \hline
    \end{tabular}
    \end{adjustbox}
\label{tab:result7}
\end{table}
\begin{table}[!htb]
    \centering
    \caption{Abalation study on different parameter setting}
    \begin{tabular}{c|c|c|c|c}
    \hline
         \multicolumn{2}{c|}{\textbf{Parameter settings}} & \multirow{2}{*}{\textbf{Top 1}} & \multirow{2}{*}{\textbf{AP50}} & \multirow{2}{*}{\textbf{DSC}} \\ \cline{1-2}
       $\lambda$ & $\tau$ & ~ & ~ & ~ \\ \hline
       0.1 & 0.2 & 0.774 & 0.9098 & 0.7994 \\ \hline
        0.25 & 0.2 & 0.7858 & 0.9138 & 0.8137 \\ \hline
         0.5 & 0.2 & 0.7894 &  0.921& 0.8219 \\ \hline
        1 & 0.2 & 0.7769 & 0.9133 & 0.8182 \\ \hline
       0.1 & 0.4 & 0.7924 & 0.9105 & 0.8092 \\ \hline
        0.25 & 0.4 & 0.7948 & 0.9198 & 0.8152 \\ \hline
         0.5 & 0.4 & \textbf{0.7977} & 0.9206 &\textbf{0.8232}  \\ \hline
         1 & 0.4 & 0.7918 & 0.9129 & 0.8207 \\ \hline
         0.1 & 0.6 & 0.7894 & 0.9045 & 0.7931 \\ \hline
       0.25 & 0.6 & 0.7918 &  0.9189& 0.817 \\ \hline
        0.5 & 0.6 & 0.7906 & \textbf{0.9227} & 0.8121 \\ \hline
       1 & 0.6 & 0.7871 & 0.9037 & 0.8039 \\ \hline
    \end{tabular}
\label{tab:result8}
\end{table}

%
\subsubsection{Loss functions}
Table~\ref{tab:result7} shows the quantitative results of our ablation study in loss functions. Initially, our proposed method, which contains three loss functions, achieves 79.12\% on top 1 accuracy and 72.09\% on the F1 score for the classification task. Similarly, it has the best AP50 and mAP of 91.92\% and 87.09\%, respectively. On the segmentation task, the combined loss also showed improvement when combined with various strategies yielding 81.13\% on DSC and 92.18\% on PPV. It can be observed that compared with classically using noise contrastive loss only, our approach and modifications led to significant improvements in all downstream tasks by a larger margin (top 1 accuracy, mAP, and DSC improved respectively by 2.61\%, 0.97\% and 2.07\%).
\subsubsection{Impact of hyper-parameters\label{sec:ablation}}
The quantitative results for the ablation study of different parameter settings are shown in Table~\ref{tab:result8}. We set different weight and temperature in Eq.~(\ref{eq:NCE_total}-\ref{eq:GCLD_loss}). Weight parameter $\lambda = \{0.1, 0.25, 0.5, 1\}$ and temperature $\tau = \{0.2, 0.4, 0.6\}$ are used for searching best parameters experimentally. As shown in Table~\ref{tab:result8}, when weight and temperature parameters are 0.5 and 0.4, respectively, our method achieves the best results in classification and segmentation tasks with 79.77\% on Top 1 accuracy and 82.32\% on DSC, respectively. For the detection task, the best performance of our SSL-CPCD was obtained when $\lambda=0.5$ and $\tau =0.6$. 
\begin{figure}[!htb]
    \centering
    \includegraphics[width=0.5\textwidth]{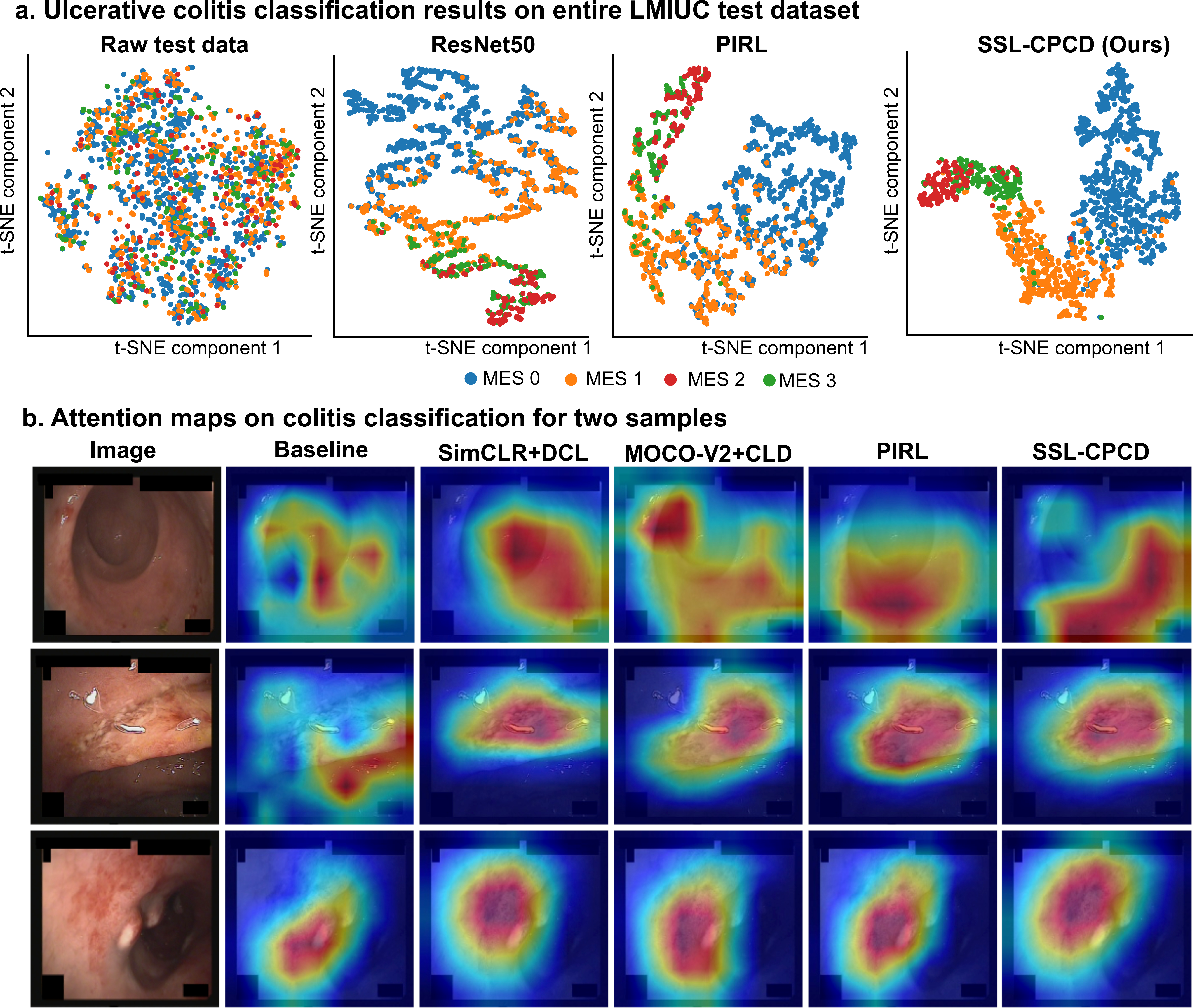}
    \caption{t-SNE plot for the raw test data, baseline network (supervised) and two SSL approaches (top), and attention maps of the proposed SSL-CPCD compared to other SOTA methods for multi-class ulcerative colitis classification task (MES 0, MES 1, and MES 2 (bottom)).}
    \label{fig:qualitative_classification}
\end{figure}
\subsection{Qualitative Analysis}
\begin{figure}[!t]
    \centering
    \includegraphics[width=0.5\textwidth]{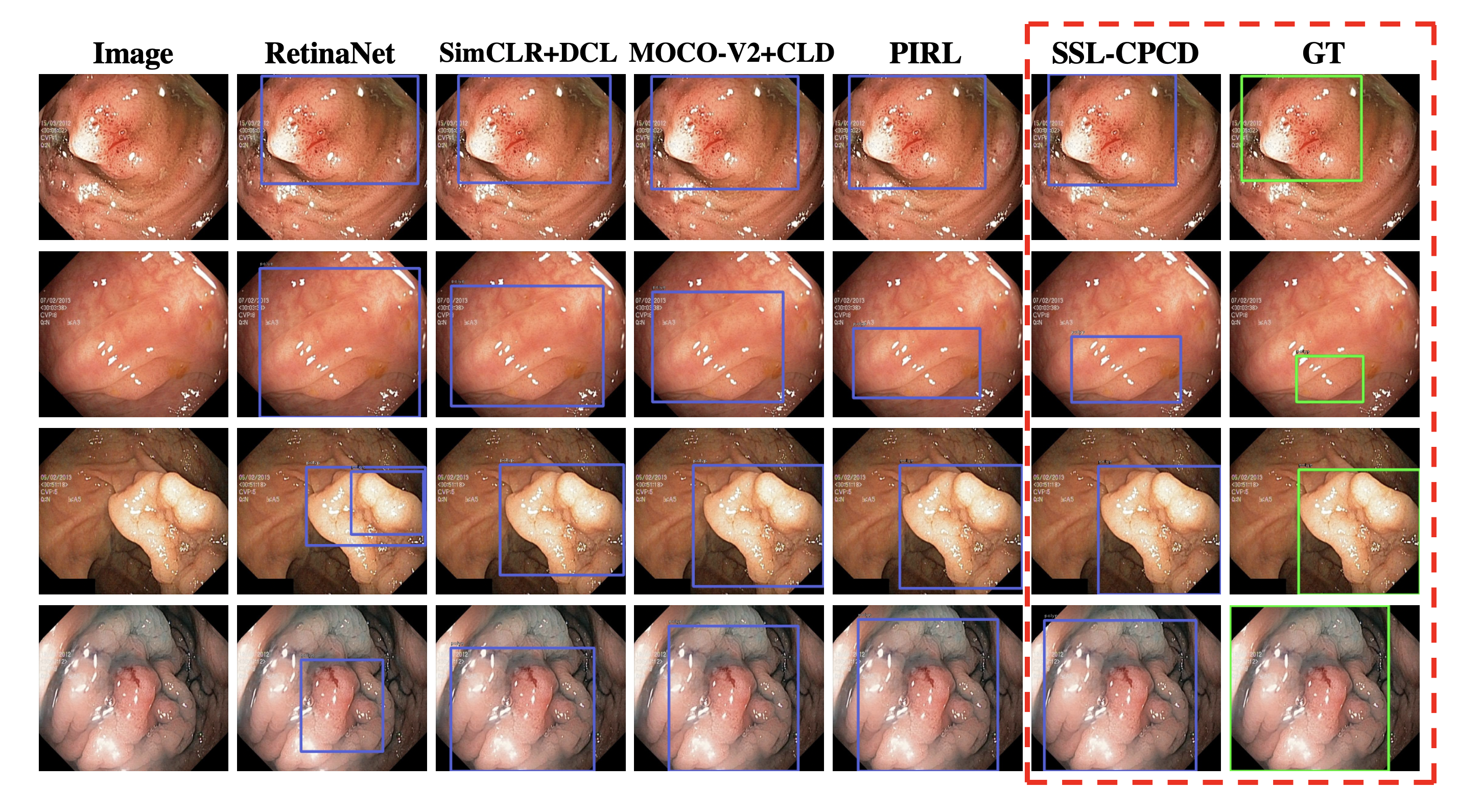}
    \caption{Qualitative comparison
    of our proposed SSL-CPCD with other SOTA methods for polyp detection task.}
    \label{fig:qualitative_detection}
\end{figure}

In the UC classification task in Fig.~\ref{fig:qualitative_classification}, a t-distributed stochastic neighbour embedding (t-SNE) plot of test image samples embedding, and gradient weighted activation map (Grad-CAM) method is used to visualise model performance. 
It can be observed (Fig.~\ref{fig:qualitative_classification} a) that the test images are stochastically distributed in raw sample distributions. After model training, images of the same class cluster in the same region. The SSL-CPCD method using clustering loss has improved more than the baseline supervised model and SSL-based PIRL approach. Using SSL-CPCD, it can be observed that the same categories are more concentrated in the same area, and there are clear boundaries between different categories, which in other cases are not apparent.
Similarly, while looking at the attention (Fig.~\ref{fig:qualitative_classification} b), the baseline method focuses on the wrong location in some images (see the first and second rows in Fig.~\ref{fig:qualitative_classification} b). In other SOTA SSL methods, the model notices the correct location, but the lesion location is not accurate. Our proposed SSL-CPCD can accurately identify the severely affected lesion area and shape. For the polyp detection task (Fig.~\ref{fig:qualitative_detection}), it can be clearly seen that baseline and other SSL methods cannot accurately locate the polyp's spatial location. Especially for the second and fourth examples in the figure, most methods have enlarged boundaries and even multiple bounding boxes. However, our SSL-CPCD approach can locate the polyp position more accurately, and the bounding boxes are closer to ground truth.

\begin{figure}[!t]
    \centering
    \includegraphics[width=0.5\textwidth]{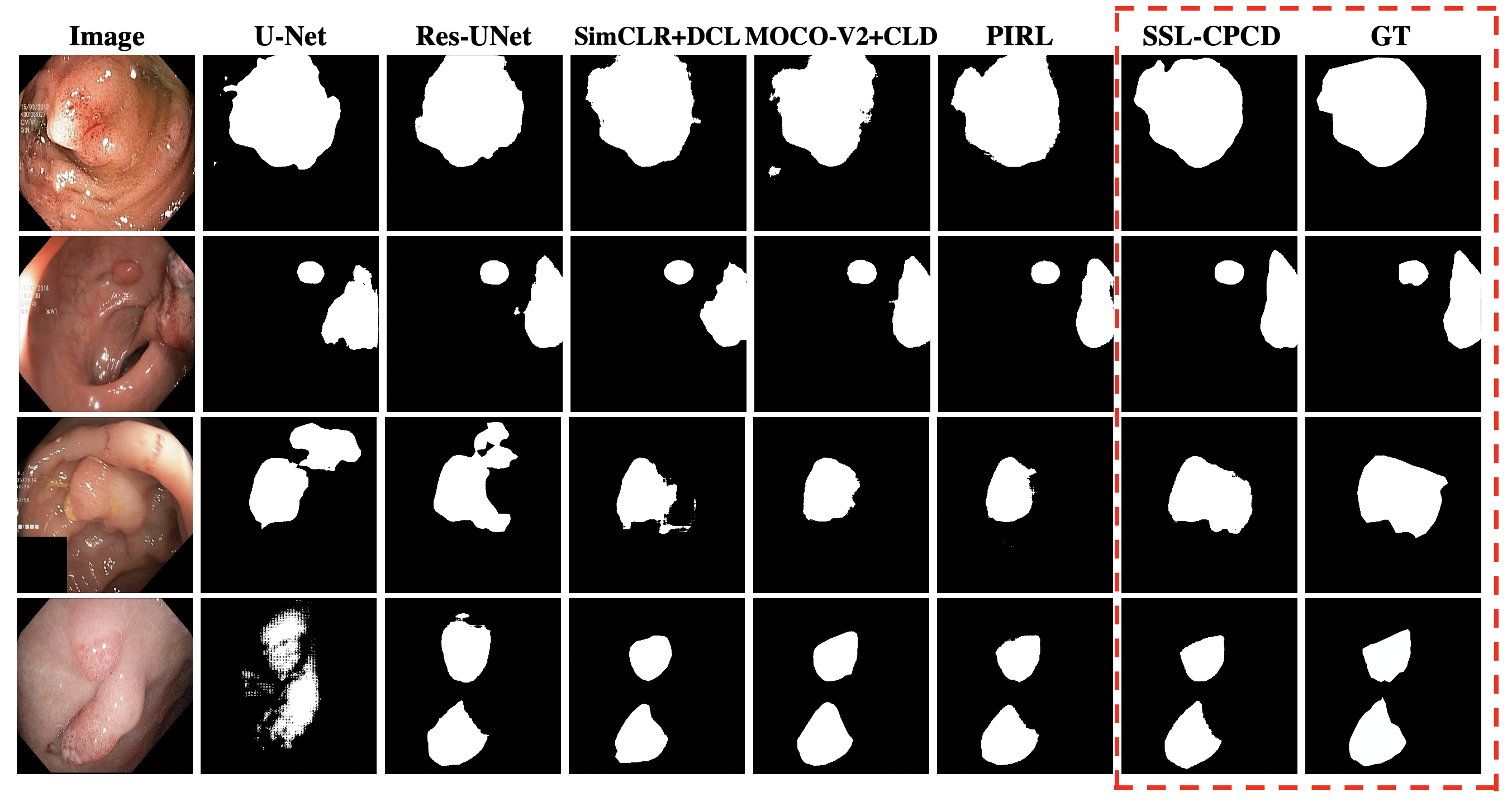}
    \caption{Qualitative comparison
    of our proposed SSL-CPCD with other SOTA methods for polyp segmentation task.}
    \label{fig:qualitative_segmentation}
\end{figure}

In the polyp segmentation task (Fig.~\ref{fig:qualitative_segmentation}), the baseline method incorrectly identifies non-polyp regions as polyps and over or under-segments the area. Although other SSL methods did not misidentify the polyp region, they only segmented part of the polyp. SSL-CPCD can segment polyps more accurately, similar to ground truth labels. Our proposed SSL-CPCD maintains the best segmentation results in all examples.

\section{Discussion and conclusion}
Supervised learning methods are unsuitable for discriminating between disease-relevant changes in the tissue structure especially in endoscopic image analysis affecting model accuracy and its generalisability capability. We explored self-supervised-based learning approach (SSL) that can learn semantically meaningful features and representations invariant to texture and illumination changes in endoscopic images that are more robust. We show that these representations using number of unlabeled endoscopic images, mitigates the risk of limited labels and provide improved results compared to widely used supervised techniques. Even though the SSL-based approaches have been proposed in the past for natural scenes ~\cite{chen2020simple,he2020momentum,misra2020self}, to our knowledge no study has been conducted comprehensively for endoscopic image analysis. We propose a novel composite pretext-class discrimination loss (CPCD) that combines noise contrastive losses for the single instance level and group-based instance showing significant improvements compared to other SSL methods. Here, instance discrimination obtains meaningful representations through instance-level contrastive learning, which can be used to reflect the apparent similarities between instances. 

The assumption that instance discrimination is established is based on the fact that each example is significantly different from others and can be treated as a separate category. However, in endoscopic image data tend to have higher similarity in their video images making it extremely hard to learn reliable features. Thus, there is a significant similarity between training data in conventional self-supervised learning, which will lead to the negative pairs used in the contrastive learning process being likely to be composed of high similarity instances, which will lead to a large number of false positives in the training process of contrastive learning repulsion. We solve this problem in two directions. First, we propose a patch-level instance-group discrimination, GCLD loss, which can perform $k$-means clustering on instances so that similar instances are clustered into the same group. The error rejection of high-similarity instances was alleviated in the subsequent contrastive loss. In addition, we further optimise the loss function by adding an angular margin $m$ between positive and negative samples in contrastive learning (see ablation study results in Table~\ref{tab:result7}). Our proposed SSL-CPCD achieves significant improvement in all three representative tasks for anomalies in colonoscopy images. In the ulcerative colitis classification task, SSL-CPCD succeeded with the highest Top 1 accuracy of 79.77\% and the highest F1 score of 72.79\% on LIMUC (see Table~\ref{tab:result1}). Likewise, we reported the highest values of 88.62\%, 94.69\%, and 92.27\% for mAP, AP25, and AP50 on Kvasir-SEG in the polyp detection task (see Table~\ref{tab:result4}). Furthermore, we report the best DSC, recall and PPV for the polyp segmentation task on the Kvasir-SEG dataset (see Table Table~\ref{tab:result5}). Furthermore, SSL-CPCD on the generalisability assessment it achieves the highest Top 1 accuracy and QWK of 67.33\% and 78.87\% (see Table~\ref{tab:result2}), and highest DSC of 67.93\% (see Table~\ref{tab:result6}).


%
To the best of our knowledge our proposed approach combining  image-level and group-level instances in a contrastive loss-based framework for self-supervised learning in endoscopic image analysis in unique and not been explored before. Our SSL-CPCD approach is capable of learning representative features from unlabeled images that are evident to improve any downstream tasks. Our strategy of added angular margin increases geometric distance between positive and negative samples. Our experiments demonstrate the effectiveness and improvement of our SSL-CPCD method over several SOTA self-supervised methods on three downstream tasks for complex colonoscopic images. Cross-dataset testing confirmed the generalisation ability of our SSL-CPCD approach which is more superior to all SOTA SSL-based methods.

\bibliographystyle{IEEEtran}
\bibliography{SSLCPCD23}

\begin{thebibliography}{10}
\providecommand{\url}[1]{#1}
\csname url@samestyle\endcsname
\providecommand{\newblock}{\relax}
\providecommand{\bibinfo}[2]{#2}
\providecommand{\BIBentrySTDinterwordspacing}{\spaceskip=0pt\relax}
\providecommand{\BIBentryALTinterwordstretchfactor}{4}
\providecommand{\BIBentryALTinterwordspacing}{\spaceskip=\fontdimen2\font plus
\BIBentryALTinterwordstretchfactor\fontdimen3\font minus
  \fontdimen4\font\relax}
\providecommand{\BIBforeignlanguage}[2]{{%
\expandafter\ifx\csname l@#1\endcsname\relax
\typeout{** WARNING: IEEEtran.bst: No hyphenation pattern has been}%
\typeout{** loaded for the language `#1'. Using the pattern for}%
\typeout{** the default language instead.}%
\else
\language=\csname l@#1\endcsname
\fi
#2}}
\providecommand{\BIBdecl}{\relax}
\BIBdecl

\bibitem{LITJENS201760}
G.~Litjens \emph{et~al.}, ``A survey on deep learning in medical image
  analysis,'' \emph{Medical Image Analysis}, vol.~42, pp. 60--88, 2017.

\bibitem{kaul2019focusnet}
C.~Kaul \emph{et~al.}, ``{FocusNet:} an attention-based fully convolutional
  network for medical image segmentation,'' in \emph{2019 IEEE 16th
  international symposium on biomedical imaging (ISBI 2019)}, 2019, pp.
  455--458.

\bibitem{Azizi2021BigSM}
S.~Azizi \emph{et~al.}, ``Big self-supervised models advance medical image
  classification,'' \emph{2021 IEEE/CVF International Conference on Computer
  Vision (ICCV)}, pp. 3458--3468, 2021.

\bibitem{chen2019self}
L.~Chen \emph{et~al.}, ``Self-supervised learning for medical image analysis
  using image context restoration,'' \emph{Medical image analysis}, vol.~58, p.
  101539, 2019.

\bibitem{AliNPJ_review22}
S.~Ali, ``Where do we stand in ai for endoscopic image analysis? deciphering
  gaps and future directions,'' \emph{npj Digital Medicine}, vol.~5, no.~1, p.
  184, Dec 2022.

\bibitem{lux2013annotation}
M.~Lux and M.~Riegler, ``Annotation of endoscopic videos on mobile devices: a
  bottom-up approach,'' in \emph{Proceedings of the 4th ACM Multimedia Systems
  Conference}, 2013, pp. 141--145.

\bibitem{zhou2021preservational}
H.-Y. Zhou \emph{et~al.}, ``Preservational learning improves self-supervised
  medical image models by reconstructing diverse contexts,'' in
  \emph{Proceedings of the IEEE/CVF International Conference on Computer
  Vision}, 2021, pp. 3499--3509.

\bibitem{huang2021towards}
W.~Huang, M.~Yi, X.~Zhao, and Z.~Jiang, ``Towards the generalization of
  contrastive self-supervised learning,'' in \emph{The Eleventh International
  Conference on Learning Representations {(ICLR)}}, 2023.

\bibitem{zhuang2019self}
X.~Zhuang \emph{et~al.}, ``Self-supervised feature learning for 3d medical
  images by playing a rubik’s cube,'' in \emph{International Conference on
  Medical Image Computing and Computer-Assisted Intervention}.\hskip 1em plus
  0.5em minus 0.4em\relax Springer, 2019, pp. 420--428.

\bibitem{nguyen2020self}
X.-B. Nguyen \emph{et~al.}, ``Self-supervised learning based on spatial
  awareness for medical image analysis,'' \emph{IEEE Access}, vol.~8, pp.
  162\,973--162\,981, 2020.

\bibitem{zeng2019sese}
Z.~Zeng \emph{et~al.}, ``Sese-net: Self-supervised deep learning for
  segmentation,'' \emph{Pattern Recognition Letters}, vol. 128, pp. 23--29,
  2019.

\bibitem{ciga2022self}
O.~Ciga \emph{et~al.}, ``Self supervised contrastive learning for digital
  histopathology,'' \emph{Machine Learning with Applications}, vol.~7, p.
  100198, 2022.

\bibitem{jha2020kvasir}
D.~Jha \emph{et~al.}, ``{Kvasir-SEG}: A segmented polyp dataset,'' in
  \emph{International Conference on Multimedia Modeling}, 2020, pp. 451--462.

\bibitem{bernal2015wm}
J.~Bernal, F.~J. S{\'a}nchez, G.~Fern{\'a}ndez-Esparrach, D.~Gil,
  C.~Rodr{\'\i}guez, and F.~Vilari{\~n}o, ``Wm-dova maps for accurate polyp
  highlighting in colonoscopy: Validation vs. saliency maps from physicians,''
  \emph{Computerized medical imaging and graphics}, vol.~43, pp. 99--111, 2015.

\bibitem{polat2022labeled}
G.~Polat \emph{et~al.}, ``Labeled images for ulcerative colitis {(LIMUC)}
  dataset,'' 2022.

\bibitem{stidham2019performance}
R.~W. Stidham \emph{et~al.}, ``Performance of a deep learning model vs human
  reviewers in grading endoscopic disease severity of patients with ulcerative
  colitis,'' \emph{JAMA network open}, vol.~2, no.~5, pp. e193\,963--e193\,963,
  2019.

\bibitem{mokter2020classification}
M.~F. Mokter, J.~Oh, W.~Tavanapong, J.~Wong, and P.~C.~d. Groen,
  ``Classification of ulcerative colitis severity in colonoscopy videos using
  vascular pattern detection,'' in \emph{International Workshop on Machine
  Learning in Medical Imaging}.\hskip 1em plus 0.5em minus 0.4em\relax
  Springer, 2020, pp. 552--562.

\bibitem{ozawa2019novel}
T.~Ozawa \emph{et~al.}, ``Novel computer-assisted diagnosis system for
  endoscopic disease activity in patients with ulcerative colitis,''
  \emph{Gastrointestinal endoscopy}, vol.~89, no.~2, pp. 416--421, 2019.

\bibitem{becker2021training}
B.~G. Becker \emph{et~al.}, ``Training and deploying a deep learning model for
  endoscopic severity grading in ulcerative colitis using multicenter clinical
  trial data,'' \emph{Therapeutic advances in gastrointestinal endoscopy},
  vol.~14, 2021.

\bibitem{lee2020real}
J.~Y. Lee \emph{et~al.}, ``Real-time detection of colon polyps during
  colonoscopy using deep learning: systematic validation with four independent
  datasets,'' \emph{Scientific reports}, vol.~10, no.~1, pp. 1--9, 2020.

\bibitem{zhang2019real}
X.~Zhang \emph{et~al.}, ``Real-time gastric polyp detection using convolutional
  neural networks,'' \emph{PloS one}, vol.~14, no.~3, p. e0214133, 2019.

\bibitem{qadir2019polyp}
H.~A. Qadir \emph{et~al.}, ``Polyp detection and segmentation using mask
  {R-CNN}: Does a deeper feature extractor {CNN} always perform better?'' in
  \emph{2019 13th International Symposium on Medical Information and
  Communication Technology (ISMICT)}, 2019, pp. 1--6.

\bibitem{shin2018automatic}
Y.~Shin \emph{et~al.}, ``Automatic colon polyp detection using region based
  deep {CNN} and post learning approaches,'' \emph{IEEE Access}, vol.~6, pp.
  40\,950--40\,962, 2018.

\bibitem{zhou2019unet++}
Z.~Zhou \emph{et~al.}, ``{UNet++}: Redesigning skip connections to exploit
  multiscale features in image segmentation,'' \emph{IEEE transactions on
  medical imaging}, vol.~39, no.~6, pp. 1856--1867, 2019.

\bibitem{fan2020pranet}
D.-P. Fan, G.-P. Ji, T.~Zhou, G.~Chen, H.~Fu, J.~Shen, and L.~Shao, ``Pranet:
  Parallel reverse attention network for polyp segmentation,'' in \emph{Medical
  Image Computing and Computer Assisted Intervention--MICCAI 2020: 23rd
  International Conference, Lima, Peru, October 4--8, 2020, Proceedings, Part
  VI 23}.\hskip 1em plus 0.5em minus 0.4em\relax Springer, 2020, pp. 263--273.

\bibitem{wei2021shallow}
J.~Wei, Y.~Hu, R.~Zhang, Z.~Li, S.~K. Zhou, and S.~Cui, ``Shallow attention
  network for polyp segmentation,'' in \emph{Medical Image Computing and
  Computer Assisted Intervention--MICCAI 2021: 24th International Conference,
  Strasbourg, France, September 27--October 1, 2021, Proceedings, Part I
  24}.\hskip 1em plus 0.5em minus 0.4em\relax Springer, 2021, pp. 699--708.

\bibitem{srivastava2021msrf}
A.~Srivastava \emph{et~al.}, ``Msrf-net: A multi-scale residual fusion network
  for biomedical image segmentation,'' \emph{Journal of Biomedical and Health
  Informatics}, vol.~26, no.~5, pp. 2252--2263, 2021.

\bibitem{jaderberg2015spatial}
M.~Jaderberg \emph{et~al.}, ``Spatial transformer networks,'' \emph{Advances in
  neural information processing systems}, vol.~28, 2015.

\bibitem{sinha2020multi}
A.~Sinha and J.~Dolz, ``Multi-scale self-guided attention for medical image
  segmentation,'' \emph{IEEE journal of biomedical and health informatics},
  vol.~25, no.~1, pp. 121--130, 2020.

\bibitem{zhao2021adasan}
Q.~Zhao \emph{et~al.}, ``Adasan: Adaptive cosine similarity self-attention
  network for gastrointestinal endoscopy image classification,'' in \emph{2021
  IEEE 18th International Symposium on Biomedical Imaging (ISBI)}, 2021, pp.
  1855--1859.

\bibitem{gu2020net}
R.~Gu \emph{et~al.}, ``Ca-net: Comprehensive attention convolutional neural
  networks for explainable medical image segmentation,'' \emph{IEEE
  transactions on medical imaging}, vol.~40, no.~2, pp. 699--711, 2020.

\bibitem{chen2020simple}
T.~Chen \emph{et~al.}, ``A simple framework for contrastive learning of visual
  representations,'' in \emph{International conference on machine
  learning}.\hskip 1em plus 0.5em minus 0.4em\relax PMLR, 2020, pp. 1597--1607.

\bibitem{he2020momentum}
K.~He \emph{et~al.}, ``Momentum contrast for unsupervised visual representation
  learning,'' in \emph{Proceedings of the IEEE/CVF conference on computer
  vision and pattern recognition}, 2020, pp. 9729--9738.

\bibitem{misra2020self}
I.~Misra and L.~v.~d. Maaten, ``Self-supervised learning of pretext-invariant
  representations,'' in \emph{Proceedings of the IEEE/CVF Conference on
  Computer Vision and Pattern Recognition}, 2020, pp. 6707--6717.

\bibitem{noroozi2016unsupervised}
M.~Noroozi and P.~Favaro, ``Unsupervised learning of visual representations by
  solving jigsaw puzzles,'' in \emph{European conference on computer
  vision}.\hskip 1em plus 0.5em minus 0.4em\relax Springer, 2016, pp. 69--84.

\bibitem{gutmann2010noise}
M.~Gutmann and A.~Hyv{\"a}rinen, ``Noise-contrastive estimation: A new
  estimation principle for unnormalized statistical models,'' in
  \emph{Proceedings of the thirteenth international conference on artificial
  intelligence and statistics}.\hskip 1em plus 0.5em minus 0.4em\relax JMLR
  Workshop and Conference Proceedings, 2010, pp. 297--304.

\bibitem{pmlr-v9-gutmann10a}
M.~Gutmann and A.~Hyvärinen, ``Noise-contrastive estimation: A new estimation
  principle for unnormalized statistical models,'' in \emph{Proceedings of the
  Thirteenth International Conference on Artificial Intelligence and
  Statistics}, vol.~9, 2010, pp. 297--304.

\bibitem{misawa2021development}
M.~Misawa \emph{et~al.}, ``Development of a computer-aided detection system for
  colonoscopy and a publicly accessible large colonoscopy video database (with
  video),'' \emph{Gastrointestinal endoscopy}, vol.~93, no.~4, pp. 960--967,
  2021.

\bibitem{chuang2020debiased}
C.-Y. Chuang \emph{et~al.}, ``Debiased contrastive learning,'' \emph{Advances
  in neural information processing systems}, vol.~33, pp. 8765--8775, 2020.

\bibitem{paszke2019pytorch}
A.~Paszke \emph{et~al.}, ``Pytorch: An imperative style, high-performance deep
  learning library,'' \emph{Advances in neural information processing systems},
  vol.~32, 2019.

\bibitem{lin2017focal}
T.-Y. Lin \emph{et~al.}, ``Focal loss for dense object detection,'' in
  \emph{Proceedings of the IEEE international conference on computer vision},
  2017, pp. 2980--2988.

\bibitem{ronneberger2015u}
O.~Ronneberger \emph{et~al.}, ``{U-Net}: Convolutional networks for biomedical
  image segmentation,'' in \emph{International Conference on Medical image
  computing and computer-assisted intervention}.\hskip 1em plus 0.5em minus
  0.4em\relax Springer, 2015, pp. 234--241.

\bibitem{zhang2018road}
Z.~Zhang, Q.~Liu, and Y.~Wang, ``Road extraction by deep residual u-net,''
  \emph{IEEE Geoscience and Remote Sensing Letters}, vol.~15, no.~5, pp.
  749--753, 2018.

\end{thebibliography}
\end{document}